\documentclass[]{article}
\usepackage[a4paper]{geometry}
\usepackage{mtsummit2023}
\usepackage{graphicx}
\usepackage{times}
\usepackage{url}
\usepackage{latexsym}
\usepackage{natbib}
\usepackage{layout}
\usepackage[hang,marginal]{footmisc}

\usepackage{amsmath}
\usepackage{graphicx}
\usepackage{multirow}
\usepackage{booktabs}
\usepackage{array}
\usepackage{arydshln}
\usepackage{CJKutf8}
\usepackage{tabularx}
\usepackage{pgfplots}
\usepackage{tikz}
\usepackage{subcaption}
\usepackage{adjustbox}
\usepackage{graphics}

\definecolor{navyBlue}{HTML}{0072BD}
\definecolor{darkOrange}{HTML}{D95319}
\definecolor{darkYellow}{HTML}{EDB120}
\definecolor{smoothGreen}{HTML}{77AC30}

\definecolor{set2-1}{HTML}{8ECFC9}
\definecolor{set2-2}{HTML}{FFBE7A}
\definecolor{set2-3}{HTML}{FA7F6F}
\definecolor{set2-4}{HTML}{82B0D2}
\definecolor{set2-5}{HTML}{BEB8DC}
\definecolor{set2-6}{HTML}{E7DAD2}

\begin{document}

\title{\bf Human‑in‑the‑loop Machine Translation with Large Language Model}

\author{
\name{\bf Xinyi Yang\footnotemark[1]} \hfill \addr{nlp2ct.xinyi@gmail.com}\\
\name{\bf Runzhe Zhan\thanks{~~Equal Contribution. Xinyi Yang contributes to the experiments, data curation, and analysis. Runzhe Zhan contributes to the methodology, code skeleton, and paper drafting.}} \hfill  \addr{nlp2ct.runzhe@gmail.com} \\
\name{\bf Derek F. Wong\thanks{~~Corresponding Author.}} \hfill \addr{derekfw@um.edu.mo}\\
\name{\bf Junchao Wu} \hfill \addr{nlp2ct.junchao@gmail.com}\\
\name{\bf Lidia S. Chao} \hfill \addr{lidiasc@um.edu.mo}\\
\addr{
  NLP$^2$CT Lab, Department of Computer and Information Science, University of Macau}
}

\maketitle
\pagestyle{empty}

\begin{abstract}
The large language model (LLM) has garnered significant attention due to its in-context learning mechanisms and emergent capabilities. The research community has conducted several pilot studies to apply LLMs to machine translation tasks and evaluate their performance from diverse perspectives. However, previous research has primarily focused on the LLM itself and has not explored human intervention in the inference process of LLM. The characteristics of LLM, such as in-context learning and prompt engineering, closely mirror human cognitive abilities in language tasks, offering an intuitive solution for human-in-the-loop generation.
In this study, we propose a human-in-the-loop pipeline that guides LLMs to produce customized outputs with revision instructions. 
The pipeline initiates by prompting the LLM to produce a draft translation, followed by the utilization of automatic retrieval or human feedback as supervision signals to enhance the LLM's translation through in-context learning. 
The human-machine interactions generated in this pipeline are also stored in an external database to expand the in-context retrieval database, enabling us to leverage human supervision in an offline setting.
We evaluate the proposed pipeline using the GPT-3.5-turbo API on five domain-specific benchmarks for German-English translation. The results demonstrate the effectiveness of the pipeline in tailoring in-domain translations and improving translation performance compared to direct translation instructions. Additionally, we discuss the experimental results from the following perspectives: 1) the effectiveness of different in-context retrieval methods; 2) the construction of a retrieval database under low-resource scenarios; 3) the observed differences across selected domains; 4) the quantitative analysis of sentence-level and word-level statistics; and 5) the qualitative analysis of representative translation cases.
The code and data are available at \url{https://github.com/NLP2CT/HIL-MT/}. 

\end{abstract}

\section{Introduction}
Large language models (LLMs) have exhibited remarkable proficiency in comprehending natural language prompts \citep{OpenAI2023GPT4TR, Touvron2023LLaMAOA}, enabling them to execute various controllable generation tasks based on human instructions. Furthermore, LLMs can acquire knowledge from limited demonstrations that are relevant to the input data and generate desired outputs through analogy. This paradigm, known as in-context learning (ICL) \citep{dong2022survey}, represents a significant advancement in prompt engineering and offers insights into adapting LLMs to downstream tasks without the need for fine-tuning the models.

Machine translation (MT) serves as a representative sequence-to-sequence task that also requires tailoring models to produce domain-specific translations. Traditional approaches to building domain-specific MT models involve fine-tuning pre-trained models with domain data or utilizing domain adaptation techniques to transfer in-domain MT models to out-of-domain requirements. However, these methods are suited for accessible, medium-scale MT models, which may not be suitable for LLMs. Notably, certain LLMs available through application programming interfaces (API) lack accessible weight matrices. Furthermore, optimizing LLM parameters with domain data can be expensive under resource-limited scenarios. Consequently, current research on LLM-based MT predominantly focuses on ICL, including in-context selection methods \citep{Agrawal2022IncontextES}, in-context prompt engineering \citep{Zhang2023PromptingLL}, and the systematic evaluation of LLM-based MT \citep{Hendy2023HowGA, Jiao2023IsCA}. While these lines of research present empirical studies investigating ICL in the MT task, they still exhibit a dearth of exploration in customizing LLMs for domain-specific needs. Given the representative capabilities of LLMs, which rely on generating outputs based on provided instructions, it is intuitive to leverage LLMs to refine general MT outputs for different domains.

Nevertheless, there remain challenges and issues when utilizing ICL to adapt LLMs for domain-specific needs. Firstly, ICL demonstrations for MT typically comprise source input and target reference, lacking domain features. Secondly, the ICL retrieval database is usually constructed using separate labeled data, and the retrieved demonstrations fail to capture LLMs' translation preferences. Last but not least, the adaptation of black-box LLMs does not benefit from parameter optimization, thereby limiting adaptation methods to modifying ICL inputs alone. In response to these challenges, we propose integrating LLM-specific translation feedback into ICL inputs, enabling the model to learn from both relevant input-output pairs and domain preferences.

Specifically, the proposed pipeline consists of two essential parts: feedback collection, and in-context refinement. 
To collect LLM-specific feedback associated with domains, we first request the LLM to produce domain-specific translations and obtain feedback by comparing its translation with a reference translation. The feedback takes the form of a sequence of revision instructions, indicating the necessary edits to transform the LLM's translation into the reference translation. Ideally, these revision instructions originate from human feedback sources. However, due to resource limitations, we simulate this process and generate synthetic human feedback in this study. Subsequently, these translation texts and feedback are stored together in the ICL retrieval database. For in-context refinement, when faced with new in-domain translation requests, the pipeline initially prompts the LLM to generate a draft translation. It then retrieves similar translation pairs and their revision histories as in-context demonstrations tied to the specific domain. Finally, the model refines the draft translation based on the retrieved domain-specific demonstrations. Any new human-machine interactions generated within this pipeline are incorporated to expand the in-context retrieval database. Overall, the primary concept revolves around enabling the LLM to revise its outputs by learning from relevant domain-specific revision feedback.

We conduct experiments using the proposed pipeline in the German-English translation direction across five domains, utilizing the GPT-3.5 Turbo API as a testbed for the black-box LLM. The results demonstrate that the proposed pipeline enhances domain translation performance in selected domains, as indicated by four automated evaluation metrics. To further elucidate the effectiveness of our approach, we discuss the pipeline and results through an ablation study, quantitative analysis, and qualitative analysis.

\section{Methodology}

\begin{figure}
    \centering
    \includegraphics[width=\textwidth]{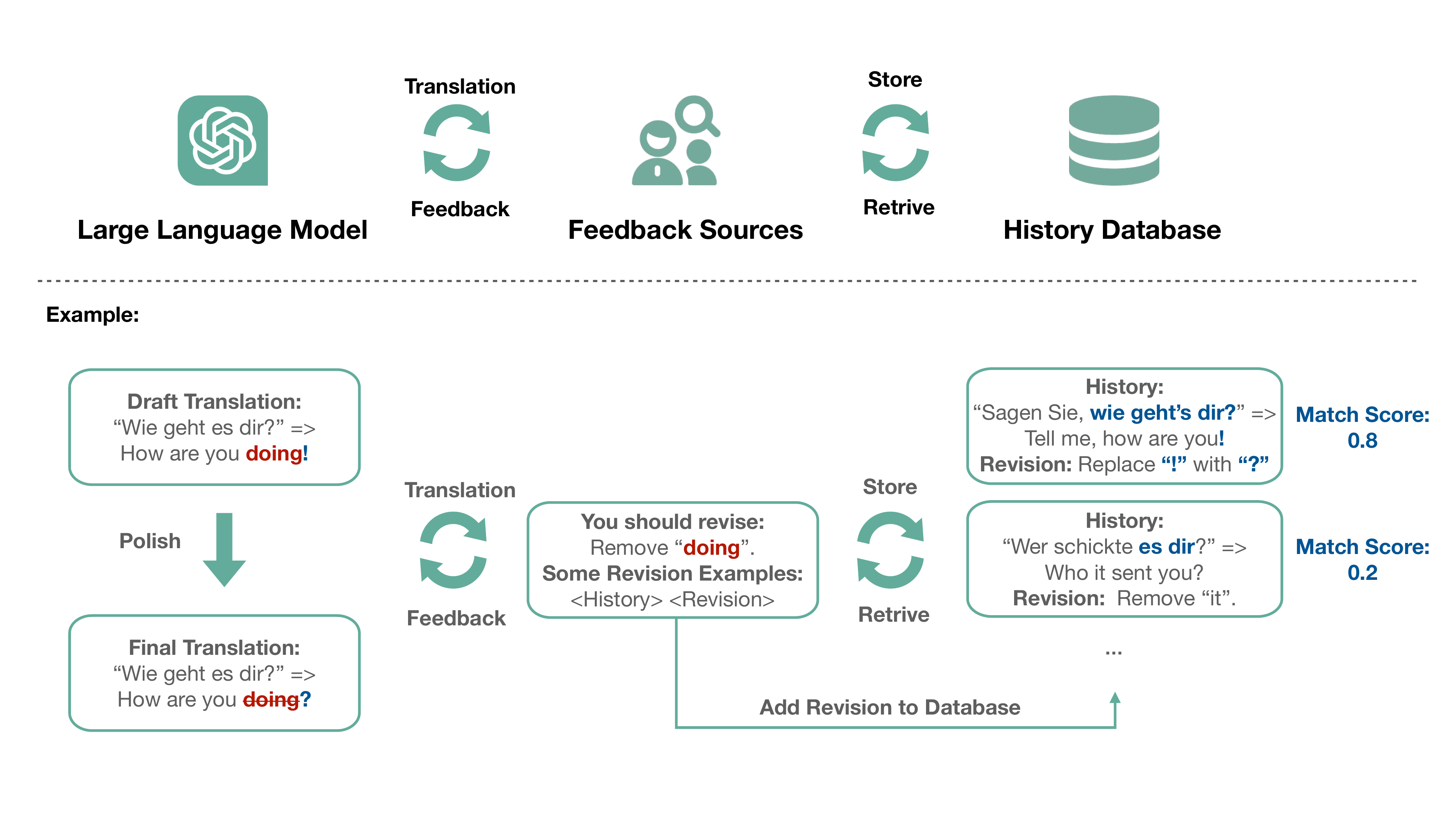}
    \caption{Illustration of the proposed human-in-the-loop translation method in the context of the large language model.}
    \label{fig:my_label}
\end{figure}
We leave the discussion of related work in the Appendix \ref{apdx} due to page limitation. The overall pipeline is illustrated in Figure \ref{fig:my_label}. 
The feedback retrieved from a data store aims to provide domain-specific revision demonstrations to LLM. Furthermore, it can also be jointly used with external human supervision in real-world applications. 
In addition to the feedback that exists in the database, any newly produced feedback for current text will be also recorded into the database to be used as a candidate for ICL retrieval in the future.

\subsection{Feedback Collection}\label{feedbackco}
To correct LLM's bias in domain translation through the ICL paradigm, we must first construct an ICL retrieval database that reflects the gap between LLM's translation preferences and domain preferences. To do this, we ask LLM to translate several domain texts first, and then use automated methods or human intervention to generate feedback on LLM's translation.
To simulate the process of human feedback, we use an automated evaluation method based on edit distance theory to generate feedback for the translations. 
Computing edit distance is a dynamic programming problem, where the cost matrix reflects what editing operations are needed to transform from the machine translation to the reference translation.
Specifically, bottom-up recursion yields the minimum cost of editing the translation, so we can generate human-like feedback by back-tracing the optimal alignment of the cost matrix and converting it to natural language. 
There are three kinds of editing operations are considered in our case: deletion, insertion, and substitution. Given an LLM's translation $h$ and reference translation $r$, the cost matrix $D(i,j)$ indicates the edit distance between $h_{<i}$ and $r_{<j}$.
Let the cost of deletion, insertion, and substitution be 1, the cost matrix $D(i,j)$ can be calculated as:
\begin{equation}
    {\begin{aligned}
    D(i,0)&=i\\
    D(0,j)&=j \\
    D(i,j)&={\begin{cases}D(i-1,j-1)&{\text{if}}\;h_{i}=r_{j}\\\min {\begin{cases}D(i-1,j)+1 \\D(i,j-1)+1\\D(i-1,j-1)+1\end{cases}}&{\text{if}}\;h_{i}\neq r_{j}\end{cases}}\end{aligned}}
\end{equation}

\begin{figure}
    \centering
    \includegraphics[width=\textwidth]{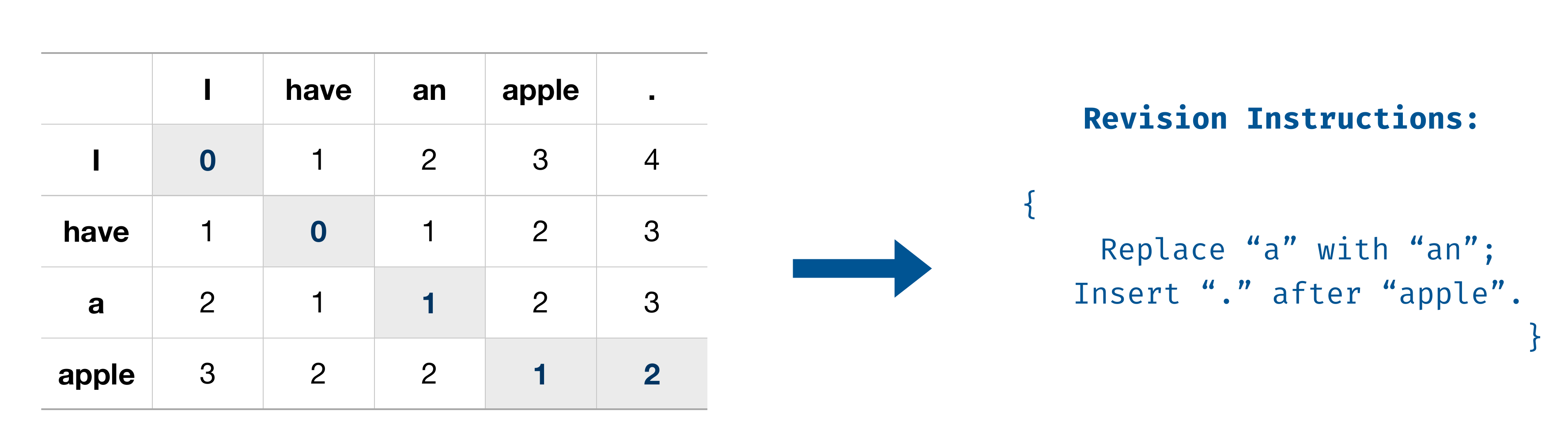}
    \caption{An example of describing the optimal path of minimum edit distance.}
    \label{fig:enter-label}
\end{figure}

We can obtain an optimal path of minimum edit distance by back-tracing the cost matrix $D$ and then convert this path to the natural language instruction as shown in Figure \ref{fig:enter-label}. 
Once the feedback of a specific test instance is produced, we combine it with the source text, LLM's translation, and reference translation to form the new ICL demonstration instance. This enables LLM to learn from its deviations from in-domain reference translations and how to polish the draft translation. 

\subsection{In-context refinement}
We construct a database for ICL retrieval using the automatic feedback generation method mentioned in the previous section. In this section, we will describe how to retrieve and utilize these ICL demonstrations as well as feedback records for conducting a two-stage translation.

\paragraph{Demonstration Retrieval}
We retrieve the relevant ICL demonstrations by evaluating the relevance between test instances and examples stored in the database. 
Specifically, we follow the previous exploratory work on ICL for MT and use two retrieval metrics: the BM25 score \citep{robertson1995okapi} and the BM25-rerank score \citep{Agrawal2022IncontextES}.
BM25 is a common retrieval metric that evaluates the relevance between the query and the documents. 
Here we consider each source sentence in the ICL database as a document and calculate the BM25 score.
BM25-Rerank, as a post-screening method for BM25, first selects the $N$ samples with the highest BM25 scores and re-scores these samples based on the n-gram recall score $R$ to select the top-$K$ samples as ICL demonstrations.
Let the input in source language be $s$, and a demonstration candidate stored in the database be $c$, then the recall score can be calculated as: 
\begin{equation}
    R = \exp{\frac{1}{n}\sum_{i=1}^{n} \log \frac{\text{Count}(\text{$i$-gram} \in s \cap c)}{\text{Count}(\text{$i$-gram} \in s)}}
\end{equation}

\paragraph{Two-stage Translation}
As mentioned in the previous section, we first ask the LLM to generate a draft translation, and then polish the draft translation by providing the ICL demonstrations, resulting in a two-stage translation process. 
In practice, we implement the two-stage translation through the multi-turn dialog feature of the GPT-3.5 API.
In addition, we empirically found that the LLM may incorrectly modify the draft translation after observing the ICL demonstrations. 
We attribute this phenomenon to the fact that the scale limitation of the retrieval database, which may lead to the inclusion of some irrelevant samples. 
Therefore, we asked the LLM further compare the polished translations with the draft translations at the second stage, and finally select the higher quality one.
It is worth noting that the LLM did not observe the reference translation of the test instance during the whole process but only compared the translation quality by means of self-reflection.

\section{Experiments}
\subsection{Data and Evaluation}
We verify the effectiveness of the proposed pipeline on a multi-domain German-English translation benchmark \citep{aharoni2020unsupervised}.
The test data keeps the same setting as the same benchmark, which involves five domains including IT, Koran, Law, Medical, and Subtitles.
We randomly sample 2,000 samples from the training set of each domain as the source of constructing the ICL retrieval database. 
To complete this process, we use GPT-3.5 API to generate the translation and employ the edit-distance-based method mentioned in Section \ref{feedbackco} to produce human-like feedback. 
We use several automated metrics to evaluate the translation quality, including BLEU \citep{papineni2002bleu}, TER \citep{snover2006study}, BERTScore \citep{zhang2019bertscore}, and COMET \citep{rei2020comet}.
BLEU and TER are the traditional metrics that evaluate the text overlap whereas the others can evaluate the semantic overlap based on neural networks.
We also found that APIs sometimes produce hallucinations or refuse to translate some sentences. To make a fair comparison, we manually check the translation results and remove the invalid results, and only evaluate the performance of the sentences that are successfully translated by all the methods.   

\subsection{Settings}
The experiments were conducted using GPT-3.5 API. The decoding temperature and the top\_p parameters are set to $1$ by default. 
When evaluating the relevance of demonstrations, we first retrieve the top $K$=200 demonstrations with the highest BM25 scores and then select top-$N$ demonstrations with the highest $4$-gram re-rank scores as the finalized ICL demonstrations.

\subsection{Main Results}
Table \ref{mainresults} presents the automated evaluation results of the baseline methods. The experimental findings demonstrate that the proposed approach effectively enhances the performance of GPT-3.5-Turbo baseline. Importantly, we observe that the impact of the proposed HIL method varies across different domains, a topic that will be thoroughly examined in Section \ref{domaind}. 
In addition, we were limited to varying the number of ICL demonstrations from 1 to 3 due to constraints on request tokens. Nonetheless, the results strongly indicate that providing more ICL demonstrations leads to improved performance.
Moreover, when evaluating the performance with neural metrics, the differences in scores are not substantial compared to the traditional metrics. We postulate that the process of refining the draft translation may not deviate significantly from the original semantic content but rather brings it closer to specific translation preferences in certain domains. These findings will be elucidated through a detailed case study in the subsequent section.

\begin{table}
\scalebox{0.9}{
\begin{tabular}{rcccccccc}
\toprule
\multicolumn{1}{l}{}   & \multicolumn{4}{c}{\textbf{IT}}                                                                             & \multicolumn{4}{c}{\textbf{Koran}}                                                                          \\
\cmidrule(lr){2-5} \cmidrule(lr){6-9} 
\multicolumn{1}{l}{}   & \multicolumn{1}{c}{\textbf{BLEU}} & \multicolumn{1}{c}{\textbf{TER}} & \multicolumn{1}{c}{\textbf{BERT-F}} & \multicolumn{1}{c}{\textbf{COMET}} & \multicolumn{1}{c}{\textbf{BLEU}} & \multicolumn{1}{c}{\textbf{TER}} & \multicolumn{1}{c}{\textbf{BERT-F}} & \multicolumn{1}{c}{\textbf{COMET}} \\
\midrule
\textbf{GPT-3.5-Turbo} & 34.4                     & 62.3                    & 93.2                       & 82.5                      & 16.2                     & \textbf{74.1}           & 90.6                       & 73.4                      \\
\textit{+1-shot HIL}   & 29.0                       & 77.3                    & 92.5                       & 81.1                      & 15.7                     & 80.8                    & 90.5                       & 72.8                      \\
\textit{+2-shot HIL}   & 33.9                     & 64.7                    & 92.9                       & 82.3                      & 15.8                     & 77.7                    & 90.6                       & 73.2                      \\
\textit{+3-shot HIL}   & 32.6                     & 68.8                    & 93.0                       & 82.2                      & 16.5                     & 76.0                      & 90.7                       & 73.6                      \\
\textit{+Compare HIL}   & \textbf{35.2}            & \textbf{61.3}           & \textbf{93.3}              & \textbf{82.8}             & \textbf{16.6}            & 74.6                    & \textbf{90.7}              & \textbf{73.8}             \\
\midrule 
\multicolumn{1}{l}{}   & \multicolumn{4}{c}
{\textbf{Law}}                                                                            & \multicolumn{4}{c}{\textbf{Medical}}                                                                        \\
\midrule
\textbf{GPT-3.5-Turbo} & 37.6                     & 54.7                    & 93.7                       & 83.8                      & 40.0                       & \textbf{59.4}           & \textbf{93.9}              & 83.4                      \\
\textit{+1-shot HIL}   & 36.2                     & 59.6                    & 93.5                       & 83.1                      & 36.6                     & 67.9                    & 93.3                       & 82.1                      \\
\textit{+2-shot HIL}   & 37.0                       & 56.6                    & 93.6                       & 83.5                      & 39.2                     & 63.0                      & 93.6                       & 83.0                      \\
\textit{+3-shot HIL}   & 36.7                     & 56.5                    & 93.6                       & 83.7                      & 38.4                     & 63.0                      & 93.7                       & 83.2                      \\
\textit{+Compare HIL}   & \textbf{37.7}            & \textbf{54.5}           & \textbf{93.8}              & \textbf{83.9}             & \textbf{40.9}            & 60.1                    & \textbf{93.9}              & \textbf{83.6}             \\
\midrule
\multicolumn{1}{l}{}   & \multicolumn{4}{c}{\textbf{Subtitles}}                                                                      & \multicolumn{1}{l}{}     & \multicolumn{1}{l}{}    & \multicolumn{1}{l}{}       & \multicolumn{1}{l}{}      \\
\midrule
\textbf{GPT-3.5-Turbo} & 27.9                     & 64.8                    & 93.0                       & 80.0                      & \multicolumn{1}{l}{}     & \multicolumn{1}{l}{}    & \multicolumn{1}{l}{}       & \multicolumn{1}{l}{}      \\
\textit{+1-shot HIL}   & 26.3                     & 69.4                    & 92.5                       & 78.8                      & \multicolumn{1}{l}{}     & \multicolumn{1}{l}{}    & \multicolumn{1}{l}{}       & \multicolumn{1}{l}{}      \\
\textit{+2-shot HIL}   & 26.4                     & 67.1                    & 92.6                       & 79.4                      & \multicolumn{1}{l}{}     & \multicolumn{1}{l}{}    & \multicolumn{1}{l}{}       & \multicolumn{1}{l}{}      \\
\textit{+3-shot HIL}   & 27.4                     & 64.6                    & 93.0                       & 79.8                      & \multicolumn{1}{l}{}     & \multicolumn{1}{l}{}    & \multicolumn{1}{l}{}       & \multicolumn{1}{l}{}      \\
\textit{+Compare HIL}   & \textbf{28.0}              & \textbf{64.1}           & \textbf{93.1}              & \textbf{80.1}             & \multicolumn{1}{l}{}     & \multicolumn{1}{l}{}    & \multicolumn{1}{l}{}       & \multicolumn{1}{l}{}     \\
\bottomrule
\end{tabular}
}
\caption{Automated evaluation results of different translation strategies. ``$K$-shot HIL'' indicates the proposed HIL method with $K$ demonstrations used. ``Compare HIL'' indicates using the comparison strategy to finalize the two-stage translation.}
\label{mainresults}
\end{table}

\section{Analysis}
\begin{table}[h]
\centering
\resizebox{\linewidth}{!}{
\setlength\tabcolsep{3pt}
\begin{tabular}{lcccccccccccccccccccc}
\toprule
\multirow{2}{*}{\textbf{Method}} & \multicolumn{4}{c}{\textbf{BM25}} & \multicolumn{4}{c}{\textbf{BM25 Re-Rank}} &  \\ 
\cmidrule(lr){2-5} \cmidrule(lr){6-9} 
& \textbf{BLEU}~ & \textbf{TER}~ & \textbf{BERT-F}~ & \textbf{COMET}~ & \textbf{BLEU}~ & \textbf{TER}~ & \textbf{BERT-F}~ & \textbf{COMET}~   \\
\midrule
\multicolumn{1}{c}{IT}        & 34.9 &62.3 & 93.1  & 82.5 &  \textbf{35.2} & \textbf{61.3} & \textbf{93.3}  & \textbf{82.8}      \\
\multicolumn{1}{c}{Koran}             & 16.2 &	\textbf{74.3}  & 90.7 & 73.7 & \textbf{16.6} & 74.6 & 90.7 & \textbf{73.8}       \\
\multicolumn{1}{c}{Law}            & \textbf{38.0} & \textbf{54.2} & \textbf{93.8} & \textbf{84.2} & 37.7 & 54.5 & 93.8 & 83.9     \\
\multicolumn{1}{c}{Medical}               & 40.6 & \textbf{58.8} & 94.0 & \textbf{83.8}  & \textbf{40.9} & 60.1 & 94.0 & 83.6  \\
\multicolumn{1}{c}{Subtitles}             & \textbf{28.2} & 64.5 & 93.0  & \textbf{80.2} & 28.0 &\textbf{64.1} &  \textbf{93.1} & 80.1      \\
\bottomrule
\end{tabular}}
\caption{Automated evaluation results for \textit{3}-shot HIL with different ICL retrieval strategies.}
\label{Single-controlE2E}
\end{table}

\subsection{Effects of ICL Retrieval Methods}
To explore the potential impact of different demonstration retrieval methods on our proposed HIL translation workflow, we conducted experiments using two strategies: BM25 and BM25 Re-rank, in a 3-shot translation scenario. The comparative results are summarized in Table \ref{Single-controlE2E}.
Based on the automated metrics, both BM25 and BM25 Re-Rank methods exhibited strengths and weaknesses in various domains. Specifically, BM25 Re-Rank slightly outperformed BM25 in terms of the BLEU metric. 
The advantage of BM25 Re-Rank was particularly evident in the IT domain, as it achieved higher scores than BM25 across all metrics. However, this conclusion was reversed in the Law domain.
The advantage of the BM25 Re-Rank strategy lies in its ability to filter out repetitive context with identical queries. Consequently, this approach selects more relevant examples related to the IT domain, leading to an improvement in the quality of the translation output within this domain.
The BM25 method for demonstration retrieval focuses on document frequency and keyword matching, making it more effective in ensuring proper usage of legal terminology.
This observation underscores the necessity of adopting different strategies for demonstration retrieval across different domains to ensure the selection of contextually relevant and domain-specific examples for the target sentences.

\subsection{Domain Differences}\label{domaind}
In general, the HIL approach exhibits superior performance compared to the GPT-3.5 API baseline across all five domains, with particularly notable advantages in the IT and Medical domains. However, the differences in performance are relatively smaller in Law and Subtitles domains. These variations can be attributed to the distinct sentence styles and structures prevalent in each domain.
Upon analyzing the translation results, it becomes evident that HIL excels in IT and Medical domains by effectively aligning terminology with the reference translations. For example, consider the phrase ``Returns a character string'' in the IT domain, HIL correctly recognizes the need to use the third-person singular form of the word ``returns'' and avoids translating ``character string'' simply as ``string''. While these words may not be technical terms, their specific usage preferences in the IT domain are crucial, and such nuances cannot be captured by the GPT-3.5 API baseline.
Conversely, in domains such as Law and Subtitles, where HIL's performance is comparatively lower, the sentences tend to adhere to specific legal clauses or follow a more colloquial and concise style. As GPT-3.5 is a multi-domain language model, it already possesses substantial knowledge related to these domains, leading to satisfactory draft translations in the initial output, thereby reducing the necessity for extensive corrections through demonstrations.
Furthermore, it is worth noting that the quality of the data within the demonstration pool may not be very high, as they were randomly sampled from the training set. This behavior could also have a negative impact on the final results.

\subsection{Quantitative Analysis}
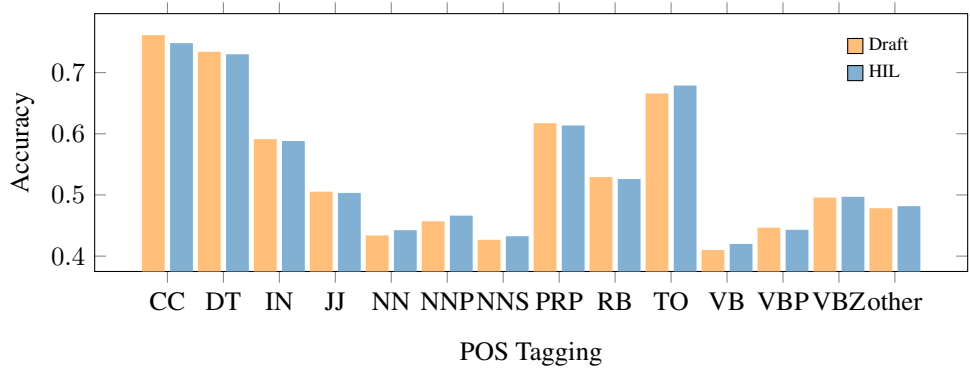
\begin{figure}
\centering
\begin{tikzpicture}
\begin{axis}[
    width=0.98\textwidth, 
    height=5cm, 
    ybar, 
    bar width=0.3cm, 
    xlabel={POS Tagging}, 
    ylabel={Accuracy}, 
    ylabel style={yshift=-8pt}, 
    xlabel style={yshift=-8pt},
    symbolic x coords={CC,DT,IN,JJ,NN,NNP,NNS,PRP,RB,TO,VB,VBP,VBZ,other}, 
    xtick=data, 
    xtick distance=1,
    enlarge x limits=0.10, 
    legend style={at={(0.90,0.95)},anchor=north, draw=none,     
    fill=none, font=\scriptsize}, 
    legend cell align={left},
    legend image code/.code={
        \draw [#1] (0cm,-0.1cm) rectangle (0.2cm,0.1cm); }, 
]

\addplot[ybar, fill=set2-2, draw=none] coordinates {
    (CC, 0.7613)
    (DT, 0.7341)
    (IN, 0.5913)
    (JJ, 0.5053)
    (NN, 0.4338)
    (NNP, 0.4569)
    (NNS, 0.4266)
    (PRP, 0.6173)
    (RB, 0.5292)
    (TO, 0.6659)
    (VB, 0.4100)
    (VBP, 0.4465)
    (VBZ, 0.4957)
    (other, 0.4783)
};

\addplot[ybar, fill=set2-4, draw=none] coordinates {
    (CC, 0.7482)
    (DT, 0.7300)
    (IN, 0.5882)
    (JJ, 0.5033)
    (NN, 0.4424)
    (NNP, 0.4662)
    (NNS, 0.4325)
    (PRP, 0.6136)
    (RB, 0.5260)
    (TO, 0.6789)
    (VB, 0.4199)
    (VBP, 0.4430)
    (VBZ, 0.4970)
    (other, 0.4817)
};

\legend{Draft, HIL}
\end{axis}
\end{tikzpicture}
\caption{The translation comparison in terms of different POS tags.}
\label{fig:POS}
\end{figure}

We conducted quantitative analysis on the translation results with the help of \texttt{compare-mt}\footnote{https://github.com/neulab/compare-mt} \citep{neubig2019compare} toolkit.
\paragraph{Part of Speech (POS)} We applied Stanford's POS tagging toolkit \citep{toutanova2003feature} to label the target-side text and subsequently examined the translation outcomes of both the baselines and HIL method across different POS categories. 
The results are presented in Figure \ref{fig:POS}.
Overall, HIL outperforms the baselines in all noun categories, including singular or mass nouns (NN), plural nouns (NNS), and singular proper nouns (NNP). Additionally, among the three verb types, HIL exhibited superior performance to the baseline in base form verbs (VB) and third person singular present verbs (VBZ), with the most noticeable advantage observed in base form verbs (VB).
Furthermore, the advantage of HIL is particularly evident in POS categories where the baseline exhibits lower translation accuracy. 
These findings underscore the effectiveness of the HIL approach in handling various parts of speech and its ability to deliver improved translation outputs, especially in challenging linguistic contexts.

\begin{figure}[h!]
  \centering
  \begin{tikzpicture}
    \begin{axis}[
      width=8cm, 
      xlabel={Sentence Length},
      ylabel={BLEU Score},
      xlabel style={yshift=-20pt},
      legend pos=north west,
      grid=both,
      xtick=data,
      xticklabels={
        $<$10,
        {[10,20)},
        {[20,30)},
        {[30,40)},
        {[40,50)},
        {[50,60)},
        $\geq$60
      },
      xticklabel style={rotate=45, anchor=near xticklabel},
      legend style={
        draw=none, 
        fill=none, 
        font=\small,
        cells={anchor=west},
        at={(0.70,0.98)}, 
        anchor=north west,
        row sep=2pt, 
      }
    ]
      \addplot table [x index=0, y index=1] {
        length Draft  Compare
        10 19.8268 20.3854
        20 30.0580 29.5361
        30 25.0412 24.7130
        40 25.3799 24.9586
        50 20.8847 20.5423
        60 27.9515 28.1203
        70 23.9318 25.3867
      };
      \addlegendentry{Draft}
      
      \addplot table [x index=0, y index=2] {
        length Draft Compare
        10 19.8268 20.3854
        20 30.0580 29.5361
        30 25.0412 24.7130
        40 25.3799 24.9586
        50 20.8847 20.5423
        60 27.9515 28.1203
        70 23.9318 25.3867
      };
      \addlegendentry{HIL}
    \end{axis}
  \end{tikzpicture}
  \caption{The translation comparison across various sentence lengths.}
  \label{fig:SENTENCE LENGTH}
\end{figure}
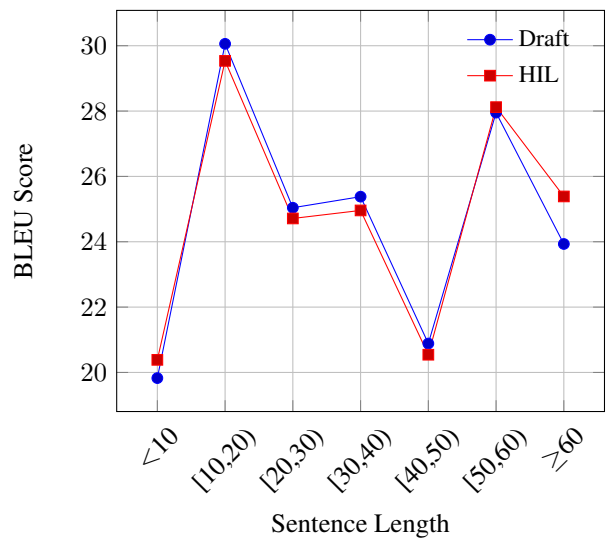

\paragraph{Sentence Length} Figure \ref{fig:SENTENCE LENGTH} presents the translation performance of both the baseline and HIL models for sentences of varying lengths. 
HIL exhibits a clear advantage over the baseline for sentences that are shorter than 10 words as well as those longer than 60 words. 
Unfortunately, HIL performs less effectively for medium-length sentences compared to the baseline's initial draft.
This observation indicates that there might be challenges specific to this sentence length range that warrant further investigation and potential refinement of the HIL approach.
\\
In conclusion, our analysis of translation performance across different POS categories and sentence lengths reveals that HIL exhibits exceptional proficiency in translating both nouns and verbs, particularly excelling in extreme-short and extreme-long sentences.

\begin{table}
    \centering
    \begin{tabular}{lp{10cm}}
        \toprule
        \multicolumn{2}{l}{IT Domain} \\
        \midrule
        Source    &  Das \textbf{Handbuch} zu \textbf{\& ksnapshot};\\
        \addlinespace[0.4\normalbaselineskip]
        Reference & The \textbf{\& ksnapshot;} \textbf{Handbook} \\
        \addlinespace[0.4\normalbaselineskip]
        Demonstrations   &  1. \textless input\textgreater Das \textbf{Handbuch zu \& kontact;} \textless hypothesis\textgreater The {manual for \& kontact;} \textless reference\textgreater \& kontact; Handbook \textless revision\textgreater ``the'' should be deleted. \underline{``manual'' should be deleted.} ``for'' should be deleted.", \underline{``handbook'' should be inserted after ``kontact;''}
        \\
       \addlinespace[0.2\normalbaselineskip]
        & 2. \textless input\textgreater Das \textbf{Handbuch} zu \& kanagram; \textless hypothesis\textgreater The \textbf{manual for \& kanagram;} \textless reference\textgreater {\& kanagram; Handbook} \textless revision\textgreater ``the'' should be deleted. \underline{``manual'' should be deleted.} ``for'' should be deleted. \underline{``handbook'' should be inserted after ``kanagramt;''} \\
       \addlinespace[0.2\normalbaselineskip]
        & 3. ... \\
        \hdashline
        \addlinespace[0.4\normalbaselineskip]
        Draft & The manual \textbf{for \& ksnapshot;}\\
        \addlinespace[0.4\normalbaselineskip]
        ICL  & The manual \textbf{for \&ksnapshot;}\\
        \addlinespace[0.4\normalbaselineskip]
        HIL  &  The \textbf{\& ksnapshot; Handbook}\\
        \bottomrule
    \end{tabular}
    \caption{An example result of three different translation strategies. ``Draft'' represents the preliminary translation results obtained at the initial turn in our HIL pipeline. ``ICL'' presents the translation results achieved using ordinary ICL demonstrations without revision feedback.}
    \label{domain example}

\end{table}

\subsection{Case Study}
Below, we present an illustrative example from the IT domain to showcase the advantages of HIL in terms of terminology translation. Additionally, we compare the proposed HIL method with the ordinary ICL method without revision feedback. The results are shown in Table \ref{domain example}.
In the given example, while the German word ``Handbuch'' was translated as ``manual'' in both the draft and ICL translations, the reference and retrieved demonstration suggest that ``Handbook'' is a more accurate translation.
Regarding the terminology usage of the IT domain, ``Handbook'' precisely describes a professional document type, whereas ``manual'' might be more generic and provide less specific information. 
Among the three translation methods, the HIL translation successfully captures this crucial information based on the provided demonstrations.
On the other hand, HIL also learns from revisions in the demonstration and opts to translate the original sentence as ``The \&ksnapshot; Handbook'', aligning more accurately with the word order of the original text. 
This demonstrates how HIL can effectively incorporate valuable revision feedback to produce more contextually appropriate and accurate translations.

\section{Conclusions and Future Work}
In this paper, we present an empirical study focused on enhancing the translation capabilities of the LLM by integrating concrete feedback within the translation process. Our objective is to establish a human-in-the-loop machine translation pipeline, where human feedback plays a pivotal role. To simulate this concept, we utilize an automated feedback method, leveraging the GPT-3.5 API as our testbed, which yields effective results.
In the future, our plan is to collect human feedback to create a novel dataset and conduct experiments on the proposed pipeline using this dataset. This approach will enable us to further validate and implement our human-in-the-loop machine translation system in the context of LLM, enhancing its practical applicability and performance.

\section*{Acknowledgement}
This work was supported in part by the Science and Technology Development Fund, Macau SAR (Grant Nos. FDCT/0070/2022/AMJ, FDCT/060/2022/AFJ), the Multi-year Research Grant from the University of Macau (Grant No. MYRG2020-00054-FST), and the Research Program of Guangdong Province (Grant No. 2220004002576). This work was performed in part at SICC which is supported by SKL-IOTSC, and HPCC supported by ICTO of the University of Macau.

\small
\bibliographystyle{apalike}
\bibliography{mtsummit2023}

\begin{thebibliography}{}

\bibitem[Agrawal et~al., 2022]{Agrawal2022IncontextES}
Agrawal, S., Zhou, C., Lewis, M., Zettlemoyer, L., and Ghazvininejad, M.
  (2022).
\newblock In-context examples selection for machine translation.
\newblock {\em ArXiv preprint}, abs/2212.02437.

\bibitem[Aharoni and Goldberg, 2020]{aharoni2020unsupervised}
Aharoni, R. and Goldberg, Y. (2020).
\newblock Unsupervised domain clusters in pretrained language models.
\newblock In {\em Proceedings of the 58th Annual Meeting of the Association for
  Computational Linguistics}, pages 7747--7763. Association for Computational
  Linguistics.

\bibitem[Dong et~al., 2023]{dong2022survey}
Dong, Q., Li, L., Dai, D., Zheng, C., Wu, Z., Chang, B., Sun, X., Xu, J., and
  Sui, Z. (2023).
\newblock A survey for in-context learning.
\newblock {\em ArXiv preprint}, abs/2301.00234.

\bibitem[Hendy et~al., 2023]{Hendy2023HowGA}
Hendy, A., Abdelrehim, M.~G., Sharaf, A., Raunak, V., Gabr, M., Matsushita, H.,
  Kim, Y.~J., Afify, M., and Awadalla, H.~H. (2023).
\newblock How good are gpt models at machine translation? a comprehensive
  evaluation.
\newblock {\em ArXiv preprint}, abs/2302.09210.

\bibitem[Jiao et~al., 2023]{Jiao2023IsCA}
Jiao, W., Wang, W., tse Huang, J., Wang, X., and Tu, Z. (2023).
\newblock Is chatgpt a good translator? yes with gpt-4 as the engine.
\newblock volume abs/2301.08745.

\bibitem[Neubig et~al., 2019]{neubig2019compare}
Neubig, G., Dou, Z.-Y., Hu, J., Michel, P., Pruthi, D., and Wang, X. (2019).
\newblock compare-mt: A tool for holistic comparison of language generation
  systems.
\newblock In {\em Proceedings of the 2019 Conference of the North {A}merican
  Chapter of the Association for Computational Linguistics (Demonstrations)},
  pages 35--41. Association for Computational Linguistics.

\bibitem[OpenAI, 2023]{OpenAI2023GPT4TR}
OpenAI (2023).
\newblock Gpt-4 technical report.
\newblock {\em ArXiv preprint}, abs/2303.08774.

\bibitem[Papineni et~al., 2002]{papineni2002bleu}
Papineni, K., Roukos, S., Ward, T., and Zhu, W.-J. (2002).
\newblock {B}leu: a method for automatic evaluation of machine translation.
\newblock In {\em Proceedings of the 40th Annual Meeting of the Association for
  Computational Linguistics}, pages 311--318. Association for Computational
  Linguistics.

\bibitem[Rei et~al., 2020]{rei2020comet}
Rei, R., Stewart, C., Farinha, A.~C., and Lavie, A. (2020).
\newblock {COMET}: A neural framework for {MT} evaluation.
\newblock In {\em Proceedings of the 2020 Conference on Empirical Methods in
  Natural Language Processing (EMNLP)}, pages 2685--2702. Association for
  Computational Linguistics.

\bibitem[Robertson et~al., 1995]{robertson1995okapi}
Robertson, S.~E., Walker, S., Jones, S., Hancock-Beaulieu, M.~M., Gatford, M.,
  et~al. (1995).
\newblock Okapi at trec-3.
\newblock {\em Nist Special Publication Sp}, 109:109.

\bibitem[Snover et~al., 2006]{snover2006study}
Snover, M., Dorr, B., Schwartz, R., Micciulla, L., and Makhoul, J. (2006).
\newblock A study of translation edit rate with targeted human annotation.
\newblock In {\em Proceedings of the 7th Conference of the Association for
  Machine Translation in the Americas: Technical Papers}, pages 223--231.
  Association for Machine Translation in the Americas.

\bibitem[Toutanova et~al., 2003]{toutanova2003feature}
Toutanova, K., Klein, D., Manning, C.~D., and Singer, Y. (2003).
\newblock Feature-rich part-of-speech tagging with a cyclic dependency network.
\newblock In {\em Proceedings of the 2003 Human Language Technology Conference
  of the North {A}merican Chapter of the Association for Computational
  Linguistics}, pages 252--259.

\bibitem[Touvron et~al., 2023]{Touvron2023LLaMAOA}
Touvron, H., Lavril, T., Izacard, G., Martinet, X., Lachaux, M.-A., Lacroix,
  T., Rozi{\`e}re, B., Goyal, N., Hambro, E., Azhar, F., Rodriguez, A., Joulin,
  A., Grave, E., and Lample, G. (2023).
\newblock Llama: Open and efficient foundation language models.
\newblock {\em ArXiv preprint}, abs/2302.13971.

\bibitem[Wang et~al., 2022]{wang2022non}
Wang, D., Wei, H., Zhang, Z., Huang, S., Xie, J., and Chen, J. (2022).
\newblock Non-parametric online learning from human feedback for neural machine
  translation.
\newblock In {\em Thirty-Sixth {AAAI} Conference on Artificial Intelligence,
  {AAAI} 2022, Thirty-Fourth Conference on Innovative Applications of
  Artificial Intelligence, {IAAI} 2022, The Twelveth Symposium on Educational
  Advances in Artificial Intelligence, {EAAI} 2022 Virtual Event, February 22 -
  March 1, 2022}, pages 11431--11439. {AAAI} Press.

\bibitem[Wu et~al., 2022]{WU2022364}
Wu, X., Xiao, L., Sun, Y., Zhang, J., Ma, T., and He, L. (2022).
\newblock A survey of human-in-the-loop for machine learning.
\newblock {\em Future Generation Computer Systems}, 135:364--381.

\bibitem[Zhang et~al., 2023]{Zhang2023PromptingLL}
Zhang, B., Haddow, B., and Birch, A. (2023).
\newblock Prompting large language model for machine translation: A case study.
\newblock {\em ArXiv preprint}, abs/2301.07069.

\bibitem[Zhang et~al., 2020]{zhang2019bertscore}
Zhang, T., Kishore, V., Wu, F., Weinberger, K.~Q., and Artzi, Y. (2020).
\newblock Bertscore: Evaluating text generation with {BERT}.
\newblock In {\em 8th International Conference on Learning Representations,
  {ICLR} 2020, Addis Ababa, Ethiopia, April 26-30, 2020}. OpenReview.net.

\end{thebibliography}

\appendix
\section{Appendix: Related Work}\label{apdx}
\subsection{ICL-based MT}
\cite{Agrawal2022IncontextES} proposed an ICL example selection method for machine translation, aiming to explore the impact of ICL examples on translation output quality. 
To address the issues in existing BM25 retrieval methods, the proposed approach re-ranks the top 100 candidate sentences selected by the BM25 score and introduces a re-ranking score, thereby maximizing the coverage of input words.
\cite{Zhang2023PromptingLL} aims to explore different prompts in the context of LLM-based machine translation. In comparison to previous related research, the main innovation of this study lies in exploring how to design prompts for LLMs to enhance their translation capabilities from three different perspectives. Specifically, the research investigates different prompting strategies, the utilization of unlabeled data, and the flexibility of transfer learning.
These research achievements demonstrate the significant value of ICL in improving the quality of machine translation systems.

\subsection{Human-in-the-Loop MT} 
The concept of Human-in-the-Loop (HIL) \citep{WU2022364} aims to leverage user feedback for optimizing the model.
Building on this idea, \cite{wang2022non} proposes a novel non-parametric online learning method called kNN-over-kNN (KoK) that does not alter the model structure. KoK is a plug-and-play non-parametric approach that learns online based on human feedback, reducing the number of user interactions and improving machine translation model performance. 
The online learning process of KoK involves three steps: decoding, correcting, and adapting. In the decoding phase, the MT system translates the source sentence, and the output is obtained by weighting the KoK method and kNN-MT. In the correcting phase, users provide corrections of the machine-translated text, resulting in the post-edited translation. Finally, in the adapting phase, post-edited and source sentences are jointly used to expand the data repositories of token-kNN and policy-kNN, thereby optimizing the model. Through these three steps, the KoK framework can promptly influence the kNN translation model's decision-making.
It is worth noting that, as of now, the HIL method has not been applied to LLM translation. Our paper focuses on exploring the potential integration of HIL into LLM translation to further enhance its performance and capabilities.

\end{document}